\documentclass[pdflatex,sn-mathphys-num]{sn-jnl}

\usepackage{placeins}  
\usepackage{array}     
\usepackage{graphicx}%
\usepackage{multirow}%
\usepackage{amsmath,amssymb,amsfonts}%
\usepackage{amsthm}%
\usepackage[title]{appendix}%
\usepackage{xcolor}%
\usepackage{textcomp}%
\usepackage{manyfoot}%
\usepackage{booktabs}%
\usepackage{algorithm}%
\usepackage{algorithmicx}%
\usepackage{algpseudocode}%
\usepackage{listings}%
\usepackage{float}
\usepackage{subcaption}
\usepackage{orcidlink}
\usepackage{caption}
\theoremstyle{thmstyleone}%
\theoremstyle{thmstyletwo}%

\theoremstyle{thmstylethree}%

\begin{document}

    









\title[Article Title]{Multi-Modal Oral Cancer Detection Using Weighted Ensemble Convolutional Neural Networks}


\author*[1]{\fnm{Ajo Babu} \sur{George}\orcidlink{0009-0005-3026-0959}}
\email{drajo\_george@dicemed.in}

\author[2]{\fnm{Sreehari} \sur{J R}\orcidlink{0009-0004-4338-7855}}


\affil[1]{\orgname{Dicemed}, 
          \orgaddress{\city{Cuttack}, \state{Odisha}, \country{India}}}

\affil[2]{\orgdiv{Department of Computer Science and Engineering}, 
          \orgname{Sree Chitra Thirunal College of Engineering}, 
          \orgaddress{\city{Thiruvananthapuram}, \state{Kerala}, \country{India}}}

\abstract{
\subsection*{Aims}
Late diagnosis of Oral Squamous Cell Carcinoma (OSCC) contributes significantly to its high global mortality rate, with over 50\% of cases detected at advanced stages and a 5-year survival rate below 50\% according to WHO statistics. This study aims to improve early detection of OSCC by developing a multimodal deep learning framework that integrates clinical, radiological, and histopathological images using a weighted ensemble of DenseNet-121 convolutional neural networks (CNNs).

\subsection*{Material and Methods}
A retrospective study was conducted using publicly available datasets representing three distinct medical imaging modalities. Each modality-specific dataset was used to train a DenseNet-121 CNN via transfer learning. Augmentation and modality-specific preprocessing were applied to increase robustness. Predictions were fused using a validation-weighted ensemble strategy. Evaluation was performed using accuracy, precision, recall, F1-score.
\subsection*{Results}
High validation accuracy was achieved for radiological (100\%) and histopathological (95.12\%) modalities, with clinical images performing lower (63.10\%) due to visual heterogeneity. The ensemble model demonstrated improved diagnostic robustness with an overall accuracy of 84.58\% on a multimodal validation dataset of 55 samples.

\subsection*{Conclusion(s)}
The multimodal ensemble framework bridges gaps in the current diagnostic workflow by offering a non-invasive, AI-assisted triage tool that enhances early identification of high-risk lesions. It supports clinicians in decision-making, aligning with global oncology guidelines to reduce diagnostic delays and improve patient outcomes.

}

\keywords{OSCC, Oral Cancer, AI in Healthcare, CNN for Cancer Detection}

\maketitle

\section{Introduction}\label{sec1}

 Oral cancer accounts for over 377,000 new cases annually worldwide, with late diagnosis contributing to high mortality rates. While histopathology remains the gold standard, integrating clinical photographs and radiological imaging (e.g., CT scans) can enhance early detection. Recent advances in convolutional neural networks (CNNs), particularly DenseNet architectures, have shown exceptional performance in medical image analysis.\cite{bib31}.
The work explores an ensemble of DenseNet 121 models trained on heterogeneous imaging modalities to improve diagnostic reliability.

Addressing a complete clinical diagnosis of oral cancer requires strong pattern identification and feature extraction from images in each modality. An end-to-end oral cancer detection model is introduced, which weighs individual DenseNet 121 Convolutional Neural Network (CNN) models based on their validation accuracies to combine predictions from clinical, radiological, and histopathological images of a patient. Each model takes respective image of its own modality, processes the image, provides a classification indicating whether the image is cancerous or not, and finally giving a weighted classification specifying whether it is cancerous or not.

The research explores the vast possibilities of artificial intelligence in the healthcare field, especially on areas where AI can be of great assistance to the medical professionals where highly precise and accurate prediction systems are required. A system capable of detecting oral cancer using multiple image types with an accuracy of 84\% is highly reliable.
\section{Background Study}\label{sec2}

Oral cancer remains a significant global health challenge with critical implications for patient survival and quality of life. Recent advancements in artificial intelligence, particularly Convolutional Neural Networks (CNNs), have shown promising results in automating the detection and classification of oral cancers.
\subsection{Convolutional Neural Networks}
The application of deep learning for oral cancer detection has seen significant growth in recent years.\cite{bib16} This surge reflects the increasing recognition of AI's potential to enhance diagnostic accuracy and efficiency in oncology. Convolutional Neural Networks (CNNs) have emerged as a powerful deep learning architecture widely adopted for medical image analysis, including oral cancer detection. Their ability to automatically learn hierarchical spatial features from input images makes them particularly suitable for identifying complex patterns in clinical, radiological, and histopathological data.

A novel methodology combining a Convolutional Neural Network (CNN) with an Improved Tunicate Swarm Algorithm (ITSA) for enhanced oral cancer detection from photographic images.\cite{bib16}. However, the training and test sets are relatively small (total images: ~161 for training, 30 for testing), which may limit the model’s generalizability and risk overfitting. Metaheuristic optimization (like ITSA) adds computational overhead, increasing training time and resource requirements compared to standard CNN training.

Exfoliative cytology images, which are non-invasive cell samples collected from the oral cavity, were compared using MobileNetV2 and other CNN models \cite{bib32}. The major disadvantages identified are that performance and generalizability of the models depend on the size, diversity, and quality of the cytology image dataset. High accuracy on a single dataset raises concerns about overfitting, especially if external validation is not performed.

A Convolutional Neural Network (CNN) whose parameters are optimized using a Combined Seagull Optimization Algorithm by PSO algorithm to get an optimal solution. \cite{bib33}
 Even though, it's to be noted that the Metaheuristic optimization significantly increases training time and computational resource requirements compared to standard gradient-based methods. Metaheuristic algorithms can be stochastic, leading to slightly different results on different runs unless random seeds and settings are carefully controlled.

Three Convolutional Neural Network (CNN) architectures, including two based on DENSENET-121, to optimize detection performance. By fusing characteristics from different architectures, the system aims to improve early detection of oral squamous cell carcinoma (OSCC). \cite{bib34} But, training and deploying deep hybrid CNN models require significant computational power and attention to avoid overfitting.

Multiclass classification on oral squamous cell carcinoma (OSCC), oral potentially malignant disorders (OPMDs) and non-pathological images by using various neural network architectures like Densenet-169, Resnet-101, SqueezeNet and Swin-S for classification on localized lesions \cite{bib35}. Nevertheless, the AUC for object detection and lesion localization in OPMDs (0.64) is lower than that for OSCC, indicating room for improvement.

A pretrained EfficientNet-B0 as a lightweight transfer learning model on a data set of 716 clinical images for binary classification of oral lesions into benign and malignant or potentially malignant. \cite{bib36} Using clinical images as a means of fundamental decision factor on oral cancer detection rises the question of ambiguity, thereby leading to the necessity of further means of diagnosis.

Various approaches have leveraged CNNs with optimization techniques to boost performance. For instance, studies have combined CNNs with metaheuristic algorithms like Improved Tunicate Swarm Algorithm (ITSA) \cite{bib16} or Particle Swarm Optimization (PSO) \cite{bib33} to fine-tune network parameters and enhance classification results. Others have modified standard architectures—such as DenseNet-121 and EfficientNet-B0—to incorporate lesion localization and multi-class classification of pre-cancerous and cancerous conditions. \cite{bib34, bib35, bib36}

Despite their success, CNN-based methods face challenges such as limited dataset sizes, overfitting, and significant computational demands, especially when incorporating hybrid or metaheuristic models. Nonetheless, their adaptability and feature extraction capabilities continue to make CNNs a foundational tool in the development of advanced diagnostic systems for oral cancer detection.

\subsection{Identified Gaps in Literature}

While numerous deep learning approaches have shown success in oral cancer detection using individual modalities—such as photographic images, cytology, or histopathology—several critical gaps remain unaddressed. First, many existing studies rely on single-modality input, which fails to capture the complementary diagnostic cues available in a real clinical workflow involving both visual inspection and internal imaging. Few, if any, have implemented \textbf{multimodal fusion} strategies that integrate clinical photographs, radiological imaging, and histopathological slides into a single predictive framework. 

Second, most reported studies evaluate models only on isolated datasets with limited cross-validation, without considering deployment in \textbf{real-world clinical pipelines} \cite{govind2024genai} or incorporating performance metrics under multi-modal consistency. Moreover, benchmarking across modalities is inconsistent, and there's a lack of \textbf{transparent ensemble strategies} that account for modality-specific strengths and weaknesses. 

Finally, only a limited number of works provide clinical interpretability or attention heatmaps (e.g., Grad-CAM) for explainability, which is critical for clinician trust and adoption. The utility of Grad-CAM and Grad-CAM++ in providing explainable insights for oral squamous cell carcinoma detection using panoramic imaging has been demonstrated \cite{11136014}, underscoring the need for a robust, generalizable multimodal system such as the one proposed in this study.

\section{Materials and Methodology}\label{sec3}
\large Identification of oral cancer through photographs poses significant challenges. Manual extraction of features and selection from photographic images for model training are being employed mostly.\cite{bib16} Two different approaches were taken for the use case.

\subsection{Dataset}

Three imaging modalities—clinical, radiological, and histopathological images—are utilized to enhance the accuracy and robustness of oral cancer detection models~\cite{bib15}. The dataset was divided into 90\% for training and 10\% for validation. The split ensured that the model had sufficient data for learning while preserving a portion for unbiased evaluation.

Data augmentation was employed to increase the diversity of the training data and reduce the likelihood of overfitting. Clinical images were augmented by applying horizontal and vertical flips, as well as slight rotations ranging from $-11^\circ$ to $+11^\circ$. Radiological images were augmented through horizontal flipping only, in order to preserve the diagnostic structure of the anatomical features. Histopathological images underwent both horizontal and vertical flipping, which is effective due to the repetitive and symmetric nature of microscopic tissue structures. All augmentation procedures were conducted in consultation with a medical professional to ensure clinical relevance and integrity.~\cite{bib17,bib20}

Radiological images were originally in DICOM format and were converted to PNG using the \texttt{pydicom} library to facilitate compatibility with standard deep learning workflows. Clinical images were resized to a resolution of 200 $\times$ 200 pixels, whereas radiological and histopathological images were resized to the resolution of 150 $\times$ 150 pixels, which produced the optimum results. During training, a batch size of 32 was used for all image types to optimize the use of available hardware resources and improve training stability.

The combination of structured preprocessing, guided augmentation, and careful dataset partitioning formed the foundation for effective training of the DenseNet-121 models across all three imaging modalities.

\par The Table \ref{tab:Dataset Source} shows details of Image Dataset used.
\renewcommand{\arraystretch}{2}
\begin{table}[!h]
\centering
\begin{tabular}{p{3cm} p{3cm} p{4cm}}
\toprule
\textbf{Type of Image Data} & \textbf{Variant} & \textbf{Source} \\
\midrule
Clinical Data & Cancer, Normal & Oral Cancer Imaging and Clinical Dataset \cite{bib37} + Oral cancer data \cite{bib27} \\
Radiological Data & Cancer & Oral Cancer Imaging and Clinical Dataset \cite{bib37} + oralcancer \cite{bib28} \\
Radiological Data & Normal & Dental Radiography Analysis and Diagnosis Dataset \cite{bib38} \\
Histopathology Data & Cancer, Normal & Histopathologic Oral Cancer Detection using CNNs \cite{bib26} \\
\bottomrule
\end{tabular}
\caption{Dataset Source}
\label{tab:Dataset Source}
\end{table}

\subsection{Weighted DenseNet (CNN)}
A weighted ensemble classification method is proposed, wherein three independently trained DenseNet-121 convolutional neural network (CNN) models are utilized—each dedicated to one imaging modality: clinical, radiological, and histopathological~\cite{bib29}. The underlying hypothesis is that combining model predictions across different modalities enhances the overall classification reliability and diagnostic accuracy, leveraging the complementary nature of multimodal medical imaging~\cite{bib14,bib15,bib19}. Each model is specifically trained to extract discriminative features from its respective input domain, enabling modality-specific learning and prediction through transfer learning.~\cite{bib20,bib18}

\par For the clinical image modality, feature maps are first extracted using a DenseNet-121 backbone~\cite{bib18}. These feature maps are passed through a dense layer comprising 128 neurons with ReLU activation to reduce dimensionality and enable high-level abstraction~\cite{bib24}. To mitigate overfitting, a dropout layer is applied, followed by a subsequent dense layer with 32 neurons and ReLU activation~\cite{bib22}. The final prediction layer consists of 2 neurons with a sigmoid activation function to perform binary classification.

\par Radiological images follow a similar architectural pipeline. Feature maps obtained from the DenseNet-121 model are forwarded to a dense layer of 128 neurons with ReLU activation~\cite{bib24}. A dropout layer is inserted for regularization~\cite{bib22}, followed by a dense layer with 64 neurons, again using ReLU activation. The final output layer is identical, with 2 neurons and a sigmoid activation function for classification.

\par For histopathological images, the DenseNet-121 model extracts the initial features. These are processed by a dense layer with 128 neurons using ReLU activation for dimensionality reduction~\cite{bib24}. The downsampled representation is fed into a dense layer with two neurons and sigmoid activation to produce the final binary prediction.

\par All models are trained using the categorical cross-entropy loss function, and their performance is evaluated based on accuracy, precision, recall, and F1-score metrics. The Adam optimizer is employed to update network weights, as it adaptively adjusts learning rates for individual parameters, facilitating efficient convergence~\cite{bib30,bib23}.

\par Transfer learning with DenseNet-121 is employed consistently across all modalities to leverage pretrained weights and improve feature extraction performance, especially given the relatively limited dataset size~\cite{bib18}. Once all individual models are trained, their outputs are ensembled using a weighted distribution. The weighting scheme in the ensemble model is based on the validation accuracy of each modality-specific CNN. This choice is grounded in the principle that validation accuracy serves as a direct, interpretable estimate of a model's generalization ability to unseen data, making it a practical and clinically meaningful metric for performance-weighted decision fusion. Unlike metrics such as sample size, which do not account for model effectiveness, or AUC, which is more relevant in highly imbalanced datasets and threshold tuning, validation accuracy reflects end-to-end performance under the binary classification task. Moreover, accuracy remains robust and intuitive in clinical decision-making contexts where binary risk stratification (cancerous vs. non-cancerous) is the goal. By assigning higher influence to models demonstrating superior validation performance, the ensemble effectively emphasizes more reliable modalities, such as radiological and histopathological data, while still incorporating potentially useful—but less consistent—clinical images. This strategy enhances the overall robustness and reduces the impact of weaker classifiers in the final decision. Weighting by validation accuracy is justified when the main objective is to maximize correct predictions on unseen data, ensuring that the weighting scheme directly supports the model's real-world performance.\cite{bib40} 

\par
\begin{figure}[H]
    \centering
    \includegraphics[width=0.57\linewidth]{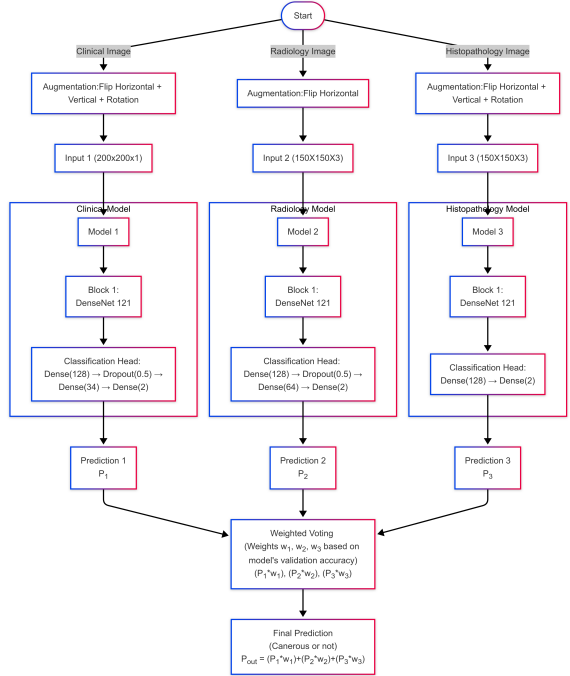}
    \caption{Architecture diagram of the weighted ensemble DenseNet-121 model}
    \label{fig:architecture}
\end{figure}

 The final ensemble prediction determines the binary classification (cancerous or non-cancerous) by aggregating the weighted outputs from the three modalities. The system is evaluated based on its collective performance across three modalities, which clearly supports the effectiveness of a multi-modality approach. The overall architecture of the proposed methodology is illustrated in Fig.\ref{fig:architecture}.

\subsection{ DenseNet Convolutional Neural Networks (CNN) }
An alternative classification strategy is proposed, wherein each DenseNet-121 model is trained independently on its respective imaging modality—clinical, radiological, and histopathological~\cite{bib29}. Unlike the ensemble-based approach discussed earlier, individual models do not aggregate predictions across modalities. Instead, it produces distinct, modality-specific outputs, with each model operating autonomously.

\par During training, feature maps are extracted from each image type using its corresponding DenseNet-121 model. These models are trained independently on their respective datasets, allowing each to specialize in recognizing features unique to its domain. Clinical images are processed exclusively by the clinical model, radiological images by the radiological model, and histopathological images by the histopathological model. As a result, each model generates its own prediction without any dependency on the outputs of the others.

\par Each model's training architecture follows the ensemble-based configuration. Feature maps are passed through dense layers for dimensionality reduction and classification, with dropout layers included to prevent overfitting. The key distinction in the method lies in the absence of any fusion or weighted combination of predictions. Each modality yields an independent classification outcome regarding the presence or absence of cancer.

\par All models are trained using the categorical cross-entropy loss function, with evaluation metrics including accuracy, precision, recall, and F1-score. The Adam optimizer is employed due to its adaptive learning rate mechanism, which facilitates efficient and stable convergence during training~\cite{bib30, bib23}.

\par Once training is complete, the performance of each model is evaluated independently based on its validation accuracy and loss. It allows for the assessment of modality-specific diagnostic effectiveness, highlighting the contribution of each image type in isolation.

\section{Results}\label{sec4}

The proposed weighted Convolutional Neural Network (CNN) architecture was evaluated using clinical, radiological, and histopathological image datasets. For clinical images, the model achieved a training accuracy, precision, and recall of 80.16\%, with an F1-score of 0.7433 and a training loss of 0.3347. However, its validation performance showed a notable decline, with an accuracy of 63.10\%, F1-score of 0.3869, and loss of 2.3356, indicating potential overfitting and increased variability in clinical data. In contrast, the model performed exceptionally well on radiological images, attaining perfect classification metrics (100\% accuracy, precision, recall, and F1-score) on both training and validation sets, with a minimal training loss of $4.41 \times 10^{-5}$ and validation loss of $1.70 \times 10^{-6}$. Histopathological image classification also yielded robust results, with 99.66\% accuracy and F1-score during training, accompanied by a low loss of 0.0118. Validation performance remained strong, recording 95.12\% accuracy, 0.9507 F1-score, and 0.1563 loss. These results highlight the model's ability to generalize effectively on radiological and histopathological data while indicating a need for better strategies to manage variability in clinical images.

\begin{figure}[htbp]
    \centering
    \includegraphics[width=0.7\linewidth]{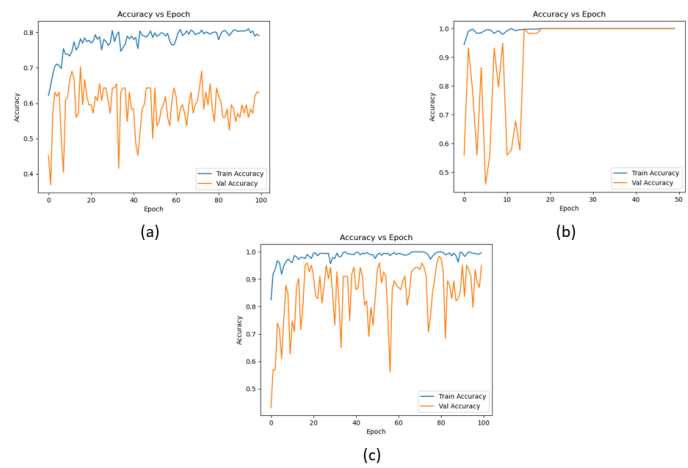}
    \caption{Model accuracies for clinical, radiological, and histopathological modalities in the order of (a), (b), and (c) respectively}
    \label{fig:accuracies}
\end{figure}

The training and testing performances of each individual model on its accuracy are plotted in Fig.\ref{fig:accuracies}
\clearpage

\section{Discussion}
\subsection{Weighted Densenet CNN}

To enhance overall classification performance and leverage the strengths of modality-specific models, a weighted ensemble approach was implemented. The final prediction was computed as a weighted sum of the outputs from three individual CNN models—trained separately on clinical, radiological, and histopathological images—where the weights were assigned based on each model’s validation accuracy, thereby distributing weights of 24.43\% for the clinical model, 38.72\% for the radiological model and 36.83\% for the histopathology model. The ensemble model outputs a binary decision: a prediction of 0 indicates a combined inference of non-cancerous (normal) across the input modalities, while a prediction of 1 denotes the presence of cancer across the input modalities.

Evaluated on a validation set comprising 55 multi-modal image samples, the weighted CNN ensemble achieved an overall accuracy of 84.58\%, with improved robustness and generalization compared to individual models. The weighted ensemble approach demonstrates the effectiveness of modality fusion in enhancing diagnostic reliability for oral cancer detection.

\subsection{Multi-Modal Deep Learning Implementation and Performance Analysis}

Transfer learning utilizing the DenseNet-121 architecture was employed to perform binary classification of oral cancer. The approach, designed to capture modality-specific features, was applied across three distinct medical imaging modalities: clinical, radiological, and histopathological. Independent models were trained for each modality. Performance assessment on the validation set revealed notable variation across modalities. Radiological and histopathological models demonstrated high generalization capabilities, achieving validation accuracies of 100\% and 95.12\%, respectively. The clinical modality model, however, showed reduced validation performance (63.10\%) compared to its training accuracy (80.16\%), attributed to higher intra-class variability and less discriminative features within this data type.

To further leverage complementary information and improve robustness, a weighted ensemble strategy was implemented. Predictions from the individual modality models were aggregated using weights derived from their respective validation accuracies. The method improved overall generalization performance, yielding a validation accuracy of 84.58\% on a validation set of 55 samples. The final ensemble model outputs a binary classification, where a positive result indicates the potential presence of cancer based on the weighted outputs of the modality-specific models. The technical effectiveness of weighted ensembling for multi-modal deep learning in scenarios with varying individual modality performance is demonstrated.
The methodology includes clearly defined index tests (modality-specific CNN models), reference standards (ground truth cancer labels), participant selection, and performance metrics (accuracy, precision, recall, F1-score) to ensure transparency, reproducibility, and clinical relevance of the reported results.

\subsection{Clinical Inference}

Integration of distinct clinical, radiological, and histopathological modalities is fundamental to a comprehensive, multi-faceted approach in oral cancer diagnosis and treatment planning. While histopathology currently serves as the diagnostic gold standard, its invasive nature can present challenges regarding patient acceptance and procedural planning. Radiological imaging, conversely, offers a non-invasive and acceptable, albeit sometimes underutilized, critical diagnostic perspective.Notably, the models demonstrated high performance on radiological (100\%) and histopathological (95.12\%) datasets, underscoring the potential of imaging data as a valuable biomarker when subjected to AI-driven analysis. Leveraging AI-based imaging assessment to complement or potentially reduce initial reliance on invasive histopathological procedures represents a potential paradigm shift. The comparatively lower accuracy observed with clinical images (63.10\%) likely reflects the inherent variability and subjectivity of visual examination compared to the more consistent patterns thdiscernible in radiology or microscopy for machine learning algorithms. A weighted ensemble strategy, integrating these diverse inputs, yielded an improved validation accuracy of 84.58\%, indicating enhanced diagnostic robustness. By suggesting the "possibility of cancer," the system functions as a valuable assistive tool for risk stratification and triage, aiding clinicians in prioritizing cases for definitive biopsy and histopathological confirmation, thereby facilitating earlier identification and intervention.

\begin{figure}[h!]
    \centering
    \includegraphics[width=0.5\linewidth]{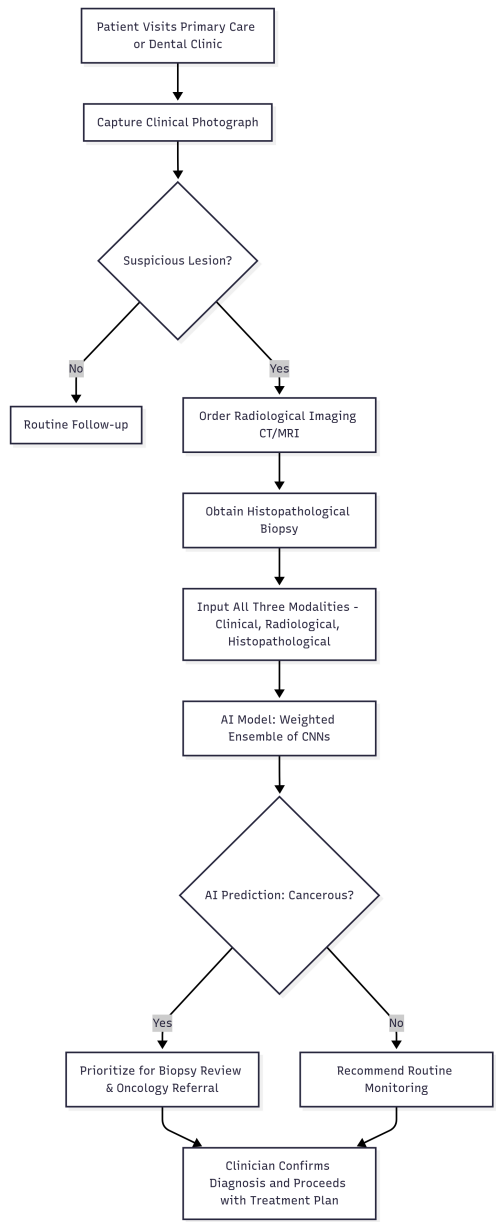} 
    \caption{Clinical workflow illustrating how the proposed AI model integrates into routine oral cancer diagnostics, supporting triage, risk stratification, and referral decisions.}
    \label{fig:clinical-workflow}
\end{figure}

By suggesting the ``possibility of cancer,'' the system functions as a valuable assistive tool for risk stratification and triage, aiding clinicians in prioritizing cases for definitive biopsy and histopathological confirmation, thereby facilitating earlier identification and intervention, as illustrated in Fig.\ref{fig:clinical-workflow}.

\clearpage

\section{Conclusion}

DenseNet layers are narrow and they introduce new feature-maps in negligible quantity, resulting to fewer parameters than conventional CNNs. The concept of feature reuse results to have fewer  parameters than other CNNs.~\cite{bib39}

A robust weighted ensembling deep learning framework is used for oral cancer detection across multiple imaging modalities. By leveraging transfer learning on the DenseNet-121 Convolutional Neural Network architecture for each modality—clinical, radiological, and histopathological—with separate classification heads~\cite{bib29}, the proposed method demonstrated high predictive reliability. Specifically, the weighted ensemble approach achieved a validation accuracy of 84.58\%, highlighting its effectiveness in aggregating complementary information from diverse data sources. 

Overall, the proposed approach demonstrates strong potential for clinical deployment, offering an accurate, multi-modal solution for early and reliable oral cancer detection.

Future extensions of this model could incorporate clinical notes and unstructured text data to further enhance diagnostic accuracy. To ensure ethical and reliable integration of such sensitive information, AI guardrail frameworks like LangChain and NeMo have been successfully applied in healthcare settings \cite{arun2025integrated}].

\section{Statements and Declarations}\label{sec7}
\noindent \textbf{Author Contributions} \newline Ajo Babu George: Conceptualization, Data Curation, Investigation, Project Administration, Resources, Writing – Review \& Editing, Supervision.
\newline
Sreehari J R: Conceptualization, Data Curation, Investigation, Methodology, Investigation, Software, Writing – Original Draft, Validation, Visualization.
\newline
All authors read and approved the submitted version.

\vspace{1em} 

\noindent \textbf{Conflict of interest} The authors declare that they have no conflicts of interest.

\vspace{1em}

\noindent \textbf{Ethical approval} This study was conducted using publicly available, de-identified datasets and did not involve any direct interaction with human participants or access to identifiable personal information. Therefore, ethical approval and informed consent were not required. The research complies with the principles outlined in the Declaration of Helsinki. All datasets used in this study are open-access and have been sourced from platforms including AIKosh, Roboflow, Zenodo, and Kaggle. 

\vspace{1em} 

\noindent \textbf{Informed consent} Informed consent was not required because the investigation did not include any human subjects.

\vspace{1em} 

\noindent \textbf{Funding} Not applicable 

\vspace{1em}


\vspace{1em}

\noindent \textbf{Limitations} This study has several limitations that must be acknowledged. First, although the model demonstrates high performance on available datasets, the total sample size—especially for multi-modality samples—is limited, with class imbalance notably affecting the clinical image subset. This raises concerns about the model's ability to generalize across broader patient populations with more heterogeneous data distributions.

Second, while care was taken to avoid data leakage within modalities, there remains a risk of subtle leakage across training and validation phases due to the absence of patient-level grouping or identifiers. Additionally, one key limitation is the \textbf{unavailability of matched patient data across all three modalities}, meaning clinical, radiological, and histopathological images used during model training and evaluation do not always originate from the same individual. This lack of multi-modality coherence at the patient level may limit the model’s effectiveness in real-world clinical deployment.

Furthermore, the model has not undergone external clinical validation or prospective evaluation in a real healthcare setting. Its current performance is based solely on retrospective data, and clinical utility in live decision support remains untested. The model’s deployment would require further testing across multiple centers and under varying acquisition conditions to verify its robustness and fairness.

Finally, while the ensemble improves overall performance by leveraging modality strengths, it may be sensitive to missing modality inputs and requires complete data for all three streams, which may not always be feasible in clinical scenarios.

\vspace{1em}

\noindent \textbf{Code Availability} The codes used for the study are available from the corresponding authors upon request.

\vspace{1em}

\noindent \textbf{Data Availability}
The datasets curated for the study are available from the corresponding authors upon request. Also,
\begin{itemize}
    \item [Dataset \citep{bib37}] AIIMS, New Delhi. Oral Cancer Imaging and Clinical Dataset. IndiaAI [AIKosh], 2024. Publicly available at \url{https://aikosh.indiaai.gov.in/home/datasets/details/oral cancer imaging and clinical dataset.}
    \item [Dataset \citep{bib27}] Vijay, Sagari. Oral cancer data Computer Vision Project. Roboflow [universe.roboflow.com], 2024.  Publicly available at \url{https://universe.roboflow.com/sagari-vijay/oral-cancer-data}
    \item [Dataset \citep{bib28}] Dental-70svr. Oral cancer data Computer Vision Project. Roboflow [universe.roboflow.com], 2024. Publicly available at \url{https://universe.roboflow.com/dental-70svr/oralcancer-blnps}
    \item [Dataset \citep{bib38}] Wang, Y., Ye, F., Chen, Y., Wang, C., Wu, C., Ma, Z., et al. STS-Tooth: A multi-modal dental dataset. Zenodo [zenodo.org], 2024. Publicly available at \url{https://zenodo.org/records/10597292}
    \item [Dataset \citep{bib26}] Kebede, Ashenafi Fasil. Histopathologic Oral Cancer Detection using CNNs dataset. Kaggle [kaggle.com], 2023.  Publicly available at \url{https://www.kaggle.com/datasets/ashenafifasilkebede/dataset/data}
    
\end{itemize}
\newpage

\newpage
\bibliography{sn-bibliography}

\end{document}